\documentclass[11pt]{article}

\usepackage[preprint]{acl}

\usepackage{times}
\usepackage{latexsym}

\usepackage[T1]{fontenc}

\usepackage[utf8]{inputenc}

\usepackage{microtype}

\usepackage{inconsolata}

\usepackage{graphicx}
\usepackage{amsmath}
\usepackage{amssymb}
\usepackage{amsfonts}

\usepackage{float}
\usepackage{hyperref}
\usepackage{url}
\usepackage{xspace}
\usepackage{booktabs}
\usepackage{multirow}
\usepackage{multicol}
\usepackage{algorithm}
\usepackage{algorithmic}
\usepackage{adjustbox}
\usepackage[most]{tcolorbox}
\usepackage{tcolorbox}
\usepackage{enumitem}
\usepackage{cuted}
\usepackage{mdframed}
\usepackage{array}
\usepackage[table]{xcolor}

\usepackage{tabularx}
\newcommand{\ourmodel}{\textsc{PaTR}\xspace}

\definecolor{codegreen}{rgb}{0,0.6,0}
\definecolor{codegray}{rgb}{0.5,0.5,0.5}
\definecolor{codepurple}{rgb}{0.58,0,0.82}
\definecolor{backcolour}{rgb}{0.95,0.95,0.92}

\lstdefinestyle{mystyle}{
    backgroundcolor=\color{backcolour},   
    commentstyle=\color{codegreen},
    stringstyle=\color{codepurple},
    basicstyle=\ttfamily\scriptsize,
    breakatwhitespace=true,         
    breaklines=true,                 
    captionpos=b,                    
    keepspaces=true,                 
    numbers=none,                    
    numbersep=5pt,                  
    showspaces=false,                
    showstringspaces=false,
    showtabs=false,                  
    tabsize=2,
    columns=flexible,
    escapeinside={(*}{*)},
}

\title{Process Reward Informed Tree Rollout for Effective Multi-Turn RL}

\author{Xintong Li$^1$\thanks{Work done during internship at Amazon.}, 
Sha Li$^2$, Yuwei Zhang$^1$, Changlong Yu$^2$, Rongmei Lin$^2$, \\
\textbf{Hongye Jin$^2$, Shuyi Guan$^3$, Xin Liu$^2$, Linwei Li$^2$, Qingyu Yin$^2$, Jingbo Shang$^1$}
\\
$^1$UC San Diego \quad 
$^2$Amazon \quad  $^3$MIT Alumni\quad \\
\texttt{\{xil240,jshang\}@ucsd.edu} \quad \texttt{slliz@amazon.com} \\}

\begin{document}
\maketitle

\begin{abstract}
Reinforcement learning (RL) has become a key approach for training LLM agents, yet popular methods such as GRPO/RLOO rely on multiple independently sampled complete trajectories for advantage estimation. 
In long-horizon agentic tasks, such a uniform rollout strategy can waste budget on uninformative dead-end attempts, while promising intermediate states do not receive sufficient exploration.
The multi-turn structure of agentic trajectories, with interleaved actions and observations, naturally supports organizing a trajectory group as a tree, where each turn serves as a decision point for exploration. This perspective reframes effective exploration as the problem of deciding where to branch.
We propose Process-Scorer Guided Adaptive Tree Rollout (\ourmodel), a quality-aware rollout framework for multi-turn agent RL. 
\ourmodel uses task-appropriate process feedback to score partial trajectories, selectively branches from promising states, reuses shared prefixes, and conservatively stops degenerate paths to reduce wasted sampling. 
The resulting rollout groups remain compatible with standard policy optimization while providing more efficient exploration under the same training budget. 
We evaluate \ourmodel on FrozenLake and the challenging SWE-Bench, which is largely unexplored by prior tree-rollout agent RL methods. 
Experiments show that \ourmodel{} improves performance by up to $+5.0$ points on SWE-Bench and $+9.3$ points on FrozenLake, highlighting process-guided tree rollouts as an effective strategy for scalable multi-turn RL.
\end{abstract}
\section{Introduction}
Large language models (LLMs) are increasingly trained as agents that solve tasks through multi-turn interactions with tools, feedback, and external observations~\citep{yao2022react,schick2023toolformer,shinn2023reflexion}.
Reinforcement learning (RL) such as GRPO~\citep{shao2024deepseekmath} has become a standard approach for improving such agentic capabilities. 
However, most existing GRPO-style agent training still relies on independently sampled complete trajectories~\citep{li2026salt}. 
This rollout strategy allocates computation uniformly across trajectories, regardless of their intermediate quality. 
For long-horizon agentic tasks, such uniform allocation can waste budget on uninformative trajectories, such as repeated tool-use loops, while promising intermediate states receive insufficient exploration.
As a result, the rollout group may contain redundant or low-value trajectories, leading to inefficient exploration and noisy advantage estimation.

\begin{figure*}[t]
    \centering
    \includegraphics[
        width=\linewidth,
        trim=0cm 1cm 0cm 0.6cm,
        clip
    ]{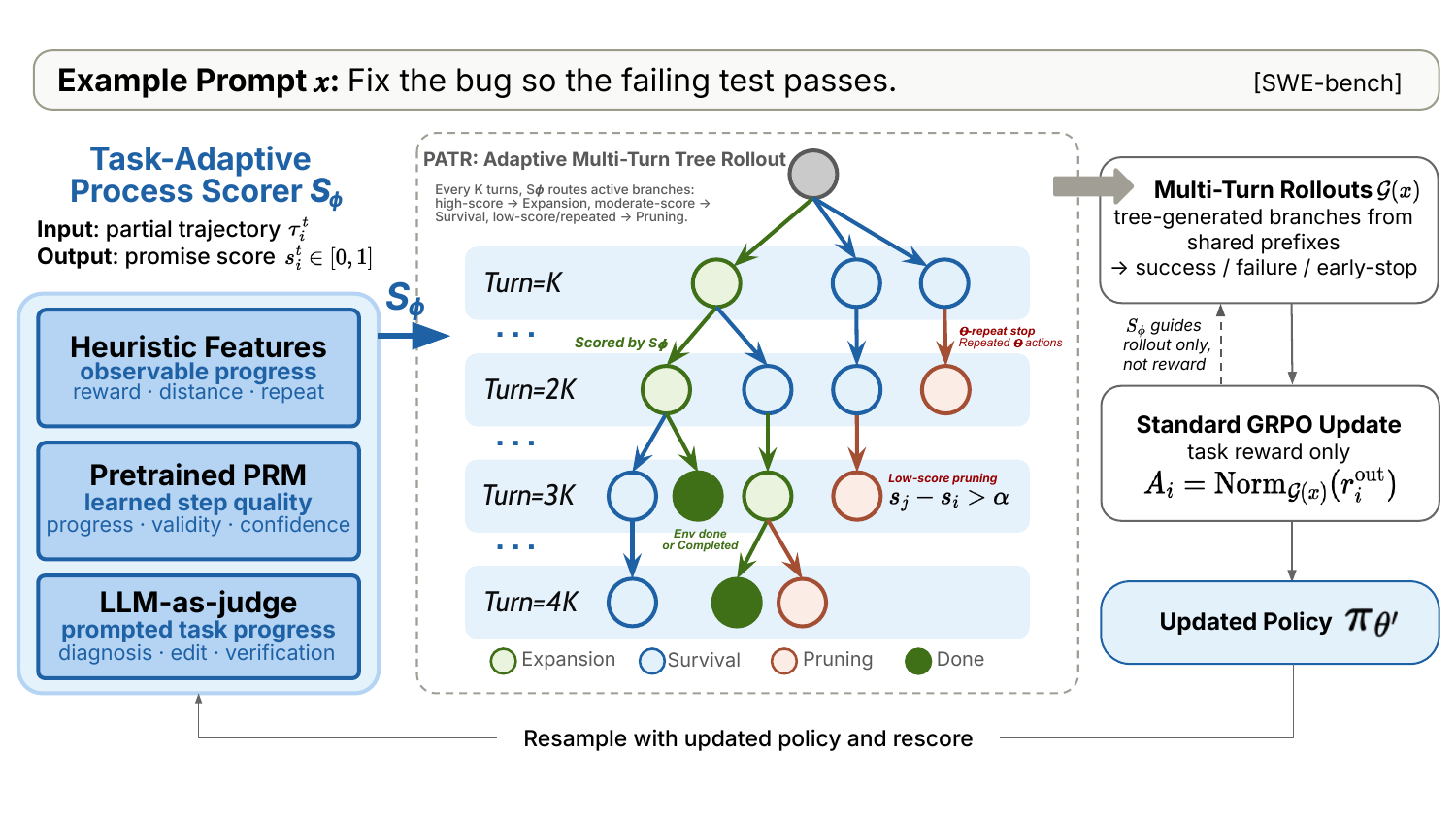}
   \caption{Overview of \ourmodel. Task-adaptive process scorer evaluates partial trajectories and guides adaptive rollout through expansion, survival, and pruning. The rollout group is optimized with standard GRPO using task rewards.}
    \label{fig:step_analysis}
    \vspace{-4mm}
\end{figure*}

Recent work has begun to improve rollout generation through tree-structured sampling, dynamic branching, and step-level feedback. 
In single-turn reasoning, tree-based methods reuse shared prefixes and explore multiple continuations from intermediate states, showing that rollout construction itself can be an important component of RL training~\citep{surana2026generatefiltercontrolreplay,xing2025lookahead}. 
This idea is even more critical for multi-turn agents, where each action changes the future context and final rewards are often sparse or delayed~\citep{feng2026group,djuhera2026tsr}.

Nevertheless, existing multi-turn tree rollout methods exhibit key limitations in how they guide exploration.
One line of work branch around high-entropy tool-use steps~\citep{dong2025agentic,dong2025agenticentropy} or allocate rollout budget using value-based uncertainty estimates~\citep{cao2026atpo}.
While these signals encourage exploration, they do not directly measure whether a partial trajectory is making meaningful progress toward the task objective.
Another line of work applies per-turn search to select the highest-scoring action at every step~\citep{djuhera2026tsr}, which can bias the rollout distribution away from the current policy and discard informative negative examples.
These limitations motivate a rollout mechanism that allocates budget based on process-level trajectory quality while preserving diverse outcomes, including failures, for policy learning.

We propose Process-Scorer Guided Adaptive Tree Rollout (\ourmodel), a cost-effective rollout generation framework for multi-turn agent RL that allocates more rollout budget to promising partial trajectories while terminating unpromising ones early.
For each training instance, \ourmodel constructs a tree of partial trajectories instead of sampling independent full trajectories from scratch.
We periodically evaluate active branches by a task-adaptive process scorer and route into one of three paths: high-scoring branches are \emph{expanded} by sampling multiple child continuations from the same intermediate state, moderate-scoring branches \emph{survive} into the next iteration, and low-scoring or degenerate branches are \emph{pruned} early.
Pruned branches are retained rather than discarded, preserving rollout diversity and serving as negative examples for policy optimization.
Since child branches inherit their parent prefixes, \ourmodel naturally reuses shared computation and produces more informative rollout trajectories under a comparable sampling budget.
We then employ standard GRPO-style policy optimization on the resulting rollout group with the same outcome rewards, avoiding direct optimization against the auxiliary process scorer.

We evaluate \ourmodel on both FrozenLake and the more challenging SWE-Bench.
FrozenLake provides a controlled testbed where process quality can be estimated with simple progress heuristics, while SWE-Bench evaluates whether process-guided tree rollouts can support realistic long-horizon coding-agent tasks.
Experiments show that \ourmodel{} improves performance by up to $+5.0$ points on SWE-Bench and $+9.3$ points on FrozenLake, highlighting the value of process-guided rollout construction for effective multi-turn RL.

Our contributions are summarized as follows:
\begin{itemize}[leftmargin=*]
    \item We propose \ourmodel, a process-guided adaptive tree rollout framework that expands promising partial trajectories, preserves surviving branches, and prunes degenerate paths.
    \item We keep \ourmodel compatible with GRPO by using process feedback only for rollout allocation while optimizing the policy with standard outcome rewards.
    \item We validate \ourmodel on FrozenLake and SWE-Bench, showing consistent gains over GRPO and other rollout-generation baselines.
\end{itemize}
\section{Related Work}
\label{sec:related}

\paragraph{Adaptive Rollout Generation for Agent RL.}
GRPO~\citep{shao2024deepseekmath} and subsequent RL-based reasoning and agent methods, such as DeepSeek-R1~\citep{guo2025deepseek} and ARPO~\citep{dong2025agentic}, commonly sample multiple trajectories per task and compute group-relative advantages from outcome rewards.
Recent work improves this rollout process through structured or adaptive sampling. 
ARPO~\citep{dong2025agentic} and AEPO~\citep{dong2025agenticentropy} branch around high-entropy tool-use steps, ATPO~\citep{cao2026atpo} allocates rollout budget using uncertainty estimates in medical dialogue, and TSR~\citep{djuhera2026tsr} applies tree-style search during training-time rollout generation.
Tree-based reasoning methods such as Tree of Thoughts~\citep{yao2023tree}, RAP~\citep{hao2023reasoning}, and tree-search agents~\citep{koh2024tree} also explore multiple continuations, but mainly for inference-time solution selection.
Our work follows the training-time rollout perspective, but uses process-level quality scores rather than entropy or uncertainty alone to guide branch expansion.

\paragraph{Process Feedback and Step-Level Supervision.}
Process reward models, introduced in step-by-step verification work such as PRM~\citep{lightman2024let}, Math-Shepherd~\citep{wang2024math}, and Skywork PRM~\citep{he2025skywork}, provide intermediate quality estimates for multi-step reasoning.
Several methods incorporate such signals into training objectives: PRIME~\citep{cui2025process} derives implicit process rewards, iStar~\citep{liu2025agentic} learns step-level rewards during agent training, WS-GRPO~\citep{mundada2026ws} constructs weakly supervised prefix rewards, and Step-GRPO~\citep{li2026step} modifies GRPO with structured process supervision.
LLM-as-judge methods~\citep{zheng2023judging} offer a flexible alternative when task-specific PRMs are unavailable, especially for open-ended agentic tasks where intermediate progress is difficult to specify manually.
In contrast, our method uses process feedback only to allocate rollout budget, leaving the policy objective and outcome reward unchanged.
\section{Preliminaries}
\label{sec:prelim}

\subsection{Multi-Turn Agentic Reinforcement Learning}
\label{sec:prelim-agent}

We consider reinforcement learning for language agents that solve tasks through multi-turn interaction with external systems.
Given a task prompt $x$, after $t$ turns the agent observes a history
$h_t=(x,a_1,o_1,\ldots,a_t,o_t)$, where $a_t$ is the agent action and $o_t$ is the corresponding observation.
The policy samples the next action as $a_{t+1}\sim\pi_\theta(\cdot\mid h_t)$.
A completed rollout $\tau=(x,a_1,o_1,\ldots,a_T,o_T)$ receives an outcome reward $r^{\mathrm{out}}(\tau)$ upon termination.
In agentic tasks, this reward is often sparse or delayed, making rollout quality critical for policy optimization~\cite{wang2025spa}.

\subsection{Group-Relative Policy Optimization}
\label{sec:prelim-grpo}

GRPO~\citep{shao2024deepseekmath} samples a group of $G$ rollouts for each task and computes group-normalized rewards, avoiding the need for a learned value function.
For trajectory $\tau_i$ in the rollout group of task $x$, the outcome advantage is,
\begin{equation}
A_i^{\mathrm{out}}
=
\frac{
r_i^{\mathrm{out}}-\mu_{\mathrm{out}}(x)
}{
\sigma_{\mathrm{out}}(x)+\varepsilon
},
\label{eq:grpo-adv}
\end{equation}
where $r_i^{\mathrm{out}}=r^{\mathrm{out}}(\tau_i)$, and $\mu_{\mathrm{out}}(x)$ and $\sigma_{\mathrm{out}}(x)$ are the mean and standard deviation of outcome rewards within the group.

The policy is optimized with the clipped importance-weighted objective
\begin{align}
\mathcal{L}(\theta)
=&\;
\mathbb{E}_{x}
\left[
\frac{1}{G}\sum_{i=1}^{G}
\frac{1}{|\tau_i|}\sum_{t=1}^{|\tau_i|}
\min\Bigl(
\rho_{i,t}(\theta)A_i^{\mathrm{out}},
\right. \notag \\
&\left.
\operatorname{clip}\bigl(\rho_{i,t}(\theta),1-\epsilon,1+\epsilon\bigr)
A_i^{\mathrm{out}}
\Bigr)
\right],
\label{eq:grpo-obj}
\end{align}
where
$\rho_{i,t}(\theta)=
\frac{\pi_\theta(a_{i,t}\mid h_{i,t-1})}
{\pi_{\theta_{\mathrm{old}}}(a_{i,t}\mid h_{i,t-1})}$
is the importance ratio.
Since advantages are computed relative to the sampled group, policy learning depends directly on rollout-group quality.
However, standard GRPO samples complete trajectories independently from the initial task state, without considering intermediate trajectory quality.
\section{\ourmodel: Process-Scorer Guided Adaptive Tree Rollout}
\label{sec:method}

We propose \textbf{\ourmodel}(\underline{P}rocess-scorer guided \underline{A}daptive \underline{T}ree \underline{R}ollout), a cost-effective rollout generation framework for multi-turn agent RL, that allocates more rollout budget to promising partial trajectories and terminate unpromising ones early, producing groups with diverse outcomes from shared prefixes.
Given a task prompt $x$, \ourmodel constructs a tree of partial trajectories and adaptively allocates rollout budget using an intermediate process scorer.
The full procedure is summarized in Algorithm~\ref{alg:patr}.

\subsection{Adaptive Tree Construction}
\label{sec:method-tree}

\ourmodel initializes each task prompt $x$ with $B_0$ active branches sampled from the current policy.
Let $\mathcal{A}$ and $\mathcal{C}$ denote the sets of active and completed branches, respectively.
At each iteration, every active branch is advanced for up to $K$ interaction steps.
Branches that terminate during this interval, such as by successful completion or reaching the maximum step budget, are moved from $\mathcal{A}$ to $\mathcal{C}$.
For each remaining active branch $\tau_i^t$ at step $t$ ($t \in \{K,2K,3K,\ldots\}$), a process scorer $S_\phi$ produces a scalar score
\begin{equation}
s_i^t = S_\phi(x,\tau_i^t) \in [0,1],
\label{eq:process-score}
\end{equation}
where higher scores indicate more promising partial trajectories for further exploration.

After scoring, each active branch is routed into one of three paths based on its process score: \textbf{expansion}, \textbf{survival}, or \textbf{pruning}.\\
\noindent\textbf{Expansion.}
High-scoring branches are selected into an expansion set $\mathcal{E}\subseteq\mathcal{A}$.
For each selected branch $\tau_i^t$, \ourmodel samples $M$ child continuations from the same intermediate state:
\begin{equation}
\tau_i^{t,m} \sim \pi_\theta(\cdot \mid x,\tau_i^t),
\qquad m=1,\ldots,M .
\label{eq:branch}
\end{equation}
These children share the parent prefix and explore different continuations, allowing \ourmodel to allocate additional samples to promising intermediate states.
The parent branch is then removed from $\mathcal{A}$ and replaced by its children.\\
\noindent\textbf{Survival.}
Branches with moderate scores remain active.
They are advanced for another $K$ steps in the next iteration and re-evaluated with updated process scores.\\
\noindent\textbf{Pruning.}
\ourmodel terminates a branch $\tau_i^t$ when its process score is substantially lower than the current active set:
\begin{equation}
s_i^t < \operatorname{median}_{j \in \mathcal{A}} s_j^t - \alpha ,
\label{eq:low-score-stop}
\end{equation}
where $\alpha$ is a margin threshold.
\ourmodel also stops deterministically degenerate branches, such as those repeating the same action for $\Theta$ consecutive turns.
Early-stopped branches are moved to $\mathcal{C}$ rather than discarded, so they remain available as negative trajectories for policy optimization.

The rollout process continues until no active branch remains.
The final rollout group is constructed from all completed trajectories.
\begin{equation}
\mathcal{G}(x)=\mathcal{C}.
\label{eq:tree-group}
\end{equation}
Importantly, process scores are used only to decide where to allocate additional rollout budget; they do not replace the task reward.
The resulting group contains both successful and failed trajectories generated from shared intermediate states, providing diverse samples for GRPO-style relative advantage estimation.

\begin{algorithm}[t]
\caption{PATR Rollout Generation}
\label{alg:patr}
\small
\begin{algorithmic}[1]
\REQUIRE Task prompt $x$; policy $\pi_\theta$; process scorer $S_\phi$;
budgets $(B_0,K,M)$; thresholds $(\alpha,\Theta)$
\STATE Initialize $\mathcal{A}$ with $B_0$ branches from $x$ under $\pi_\theta$;\quad
$\mathcal{C}\leftarrow\emptyset$
\WHILE{$\mathcal{A}\neq\emptyset$}
    \STATE Roll out each branch in $\mathcal{A}$ for up to $K$ steps under $\pi_\theta$
    \STATE Move terminated branches from $\mathcal{A}$ to $\mathcal{C}$
    \IF{$\mathcal{A}\neq\emptyset$}
        \STATE Compute $s_i^t=S_\phi(x,\tau_i^t)$ for each $\tau_i^t\in\mathcal{A}$
        \hfill Eq.~\ref{eq:process-score}
        \STATE Select high-scoring expansion set $\mathcal{E}\subseteq\mathcal{A}$
        \STATE Let $\mathcal{L}\subseteq\mathcal{A}\setminus\mathcal{E}$ be branches satisfying
        the low-score or $\Theta$-repeat criterion
        \hfill Eq.~\ref{eq:low-score-stop}
        \STATE Sample child set
        $\mathcal{B}=\{\tau_i^{t,m}:\tau_i^t\in\mathcal{E},\,m=1,\ldots,M\}$
        \hfill Eq.~\ref{eq:branch}
        \STATE $\mathcal{A}\leftarrow(\mathcal{A}\setminus(\mathcal{E}\cup\mathcal{L}))\cup\mathcal{B}$;\quad
        $\mathcal{C}\leftarrow\mathcal{C}\cup\mathcal{L}$
    \ENDIF
\ENDWHILE
\STATE \textbf{return} $\mathcal{G}(x)=\{\tau\in\mathcal{C}:|\tau|>0\}$
\end{algorithmic}
\end{algorithm}

\subsection{Task-Adaptive Process Scoring}
\label{sec:method-prm}

\ourmodel abstracts task-specific intermediate feedback into a scalar process score as defined in Eq.~\ref{eq:process-score},
This interface supports different scorer instantiations, including domain-specific heuristics, pretrained PRMs, and LLM judges, without changing the rollout algorithm or the GRPO objective.

\paragraph{Heuristic scorer.}
When intermediate progress is directly observable, we instantiate $S_\phi$ as a lightweight heuristic over the current state and accumulated step rewards.
For example, in navigation-style tasks, the scorer can combine the reward accumulated so far with a normalized progress measure such as distance to the goal.
This instantiation requires no learned model and provides a controlled setting for studying adaptive tree rollout when reliable progress signals are available.

\paragraph{Pretrained PRM scorer.}
For tasks where a pretrained process reward model is available, we convert each partial trajectory into a problem--response format.
The problem field contains the task prompt, while the response field serializes recent interaction steps, including agent actions, observations, and available step rewards.
The PRM produces step-level scores $\{c_\ell\}_{\ell=1}^{L_i}$ at designated step boundaries, which are aggregated into a branch score:
\begin{equation}
s_i^t = f_{\mathrm{agg}}(c_1,\ldots,c_{L_i}),
\label{eq:prm-agg}
\end{equation}
where $f_{\mathrm{agg}}$ can be the last-step score, the mean score, the maximum score, or a weighted combination of the last and mean scores.
For long trajectories, we truncate the serialized input while preserving the task prompt and recent interaction steps, which are most relevant for deciding whether to expand the current branch.

\paragraph{LLM-as-judge scorer.}
For complex agentic tasks where no reliable hand-crafted signal or pretrained PRM is sufficient, we instantiate $S_\phi$ with an LLM judge.
The judge receives the task prompt and a compact representation of the partial trajectory, including recent actions, observations, and automatically extracted trajectory signals.
Specifically, to make scoring sensitive to different stages of SWE-style problem solving, we use phase-specific judge prompts for diagnosis, editing, and verification. 
The judge is instructed to output a normalized JSON score indicating whether continuing from the current state is likely to lead to a successful solution.
We further blend the judge output with a lightweight heuristic based on trajectory features such as repeated actions, error or test signals in observations, evidence before editing, and targeted verification:
\begin{equation}
s_i^t = \lambda s_{\mathrm{judge}}(\tau_i^t)
+ (1-\lambda)s_{\mathrm{heur}}(\tau_i^t).
\label{eq:judge-blend}
\end{equation}
This design provides robust process scores for long-horizon tool-use tasks while keeping the interface to \ourmodel identical to the heuristic and pretrained-PRM cases.
Appendix~\ref{app:judge-prompts} provides the full prompts and describes the heuristic scoring signals in detail.

\subsection{GRPO Training with Tree-Generated Rollout Groups}
\label{sec:method-opt}

After tree rollout construction, \ourmodel uses $\mathcal{G}(x)$ from Eq.~\ref{eq:tree-group} as the rollout group for GRPO.
For each trajectory $\tau_i \in \mathcal{G}(x)$, we compute its outcome reward $r_i^{\mathrm{out}}=r^{\mathrm{out}}(\tau_i)$ and normalize rewards within the tree-generated group,
\begin{equation}
A_i^{\mathrm{tree}}
=
\frac{
r_i^{\mathrm{out}}-\mu_{\mathcal{G}}(x)
}{
\sigma_{\mathcal{G}}(x)+\varepsilon
},
\label{eq:tree-grpo-adv}
\end{equation}
where $\mu_{\mathcal{G}}(x)$ and $\sigma_{\mathcal{G}}(x)$ are computed over all valid trajectories in $\mathcal{G}(x)$.
The policy is then updated with the same clipped GRPO objective in Eq.~\ref{eq:grpo-obj}, replacing the independently sampled rollout group with the tree-generated group.

\noindent\textbf{Remark}.
\ourmodel departs from the uniform independent sampling used in vanilla GRPO by adaptively allocating more continuations to selected partial histories.
This introduces a controlled and localized bias in the rollout group.
Importantly, the process scorer is used only for branch selection: it does not alter the policy model, task reward, advantage normalization, or optimization objective.
After tree construction, completed, failed, and early-stopped branches are all retained as outcome-labeled trajectories, and the policy is updated using standard GRPO with task rewards.
This idea coincides with the efficiency-bias trade-off discussion in the literature of reinforcement learning, where non-uniform allocation of computation or samples can improve data efficiency by focusing efforts on more informative states or transitions~\cite{kocsis2006bandit,schaul2016prioritizedexperiencereplay,espeholt2018impalascalabledistributeddeeprl}.
In \ourmodel, the induced mismatch is further limited because actions are still generated by the current rollout policy conditioned on realized histories, and only the number of continuations allocated to each partial history is adapted.
Thus, \ourmodel trades strict uniform rollout sampling for more informative rollout groups while keeping the learning objective unchanged.
\section{Experiments}
\label{sec:exp}

\subsection{Experimental Setup}
\label{sec:exp-setup}

\paragraph{Tasks, datasets, and models.}
We evaluate \ourmodel on two settings with different levels of interaction complexity.
\textbf{FrozenLake}~\citep{brockman2016openai} is a grid-world navigation task in which an agent must reach a goal while avoiding holes, providing a controlled testbed for multi-turn exploration.
We use the rLLM FrozenLake environment~\citep{tan2025rllm} with procedurally generated task prompts.
\textbf{SWE-agent training} targets long-horizon software-engineering tasks, where agents must inspect repositories, edit code, and verify fixes through tool interaction.
For training, we use a filtered subset of R2E-Gym-Lite-with-Difficulty, which is derived from R2E-Gym~\citep{jain2025r2e} by removing the most difficult instances.
We evaluate on SWE-Bench Verified~\citep{jimenez2024swe}, a human-filtered set of 500 instances designed for reliable evaluation of coding agents and language models.
For FrozenLake, we train Qwen2.5-0.5B-Instruct and Qwen2.5-3B-Instruct~\citep{yang2024qwen2technicalreport}; for SWE-agent training, we use Qwen3-4B-Instruct-2507~\citep{yang2025qwen3}.

\paragraph{Baselines.}
We compare against representative GRPO-based and tree-style rollout strategies under the same policy optimization objective.
\textbf{GRPO}~\citep{shao2024deepseekmath} samples independent complete trajectories for each task.
\textbf{DAPO}~\citep{yu2026dapo} strengthens GRPO with token-level policy-gradient refinements, rejection sampling, and asymmetric clipping.
\textbf{ARPO}~\citep{dong2025agentic} performs adaptive branching for agentic rollouts based on uncertainty signals rather than process-level trajectory quality.
\textbf{Tree-Random} uses the same tree rollout structure as \ourmodel but replaces process-guided selection with random branch allocation.

\paragraph{Evaluation.}
For FrozenLake, we follow the standard success-rate evaluation. Each test prompt is sampled four times, and we report the average fraction of successful task instances.
For SWE-Bench evaluation, we use the SWE-agent scaffold with Docker-based execution environments.
Each trained agent generates one trajectory per SWE-Bench Verified instance, and we report the resolved rate, following the standard SWE-Bench protocol.

\paragraph{Implementation Details.}
All experiments are implemented with the rLLM framework~\citep{tan2025rllm}.
Across both tasks, \ourmodel starts from four initial branches and samples two child continuations when expanding a branch.
Branching is performed at fixed intervals, with $K=5$ for FrozenLake and $K=13$ for SWE-agent training.
We set the maximum interaction length to 10 turns for FrozenLake and 50 turns for SWE-agent training, with maximum response lengths of 10K and 32K tokens, respectively.
Early stopping uses the score-margin rule in Eq.~\ref{eq:low-score-stop} with $\alpha=0.5$, together with an action-loop criterion that terminates branches repeating the same action six times.

The process scorer is instantiated according to the task.
For FrozenLake, we use a lightweight progress heuristic based on accumulated rewards and shortest-path distance to the goal, treating holes as blocked cells.
For SWE-agent training, we evaluate two scorer variants: \textbf{\ourmodel-PRM}, which uses Skywork-o1-Open-PRM-Qwen-2.5-7B~\citep{he2025skywork}, and \textbf{\ourmodel-Judge}, which uses Qwen2.5-Coder-7B-Instruct~\citep{hui2024qwen2} as an LLM judge with phase-specific prompts.
Both scorers operate on recent trajectory context; additional scoring details are provided in Appendix~\ref{app:impl-details}.
To ensure a fair comparison, GRPO and DAPO use a rollout group size of 8, while ARPO, Tree-Random, and \ourmodel use the same number of initial branches.
Detailed training hyperparameters and hardware settings are reported in Appendix~\ref{app:impl-details}.

\begin{table}[t]
\centering
\small
\setlength{\tabcolsep}{7pt}
\renewcommand{\arraystretch}{1.15}
\resizebox{\linewidth}{!}{
\begin{tabular}{lcccc}
\toprule
\textbf{Method}
  & \textbf{Succ. Rate} $\uparrow$
  & \textbf{Pass@4} $\uparrow$
  & \textbf{RLen} $\downarrow$
  & \textbf{Turns} $\downarrow$ \\
\midrule
\multicolumn{5}{l}{\textit{\textbf{Qwen2.5-0.5B-Instruct}}} \\
GRPO
  & 66.5 & 67.0 & 513 & 3.4 \\
DAPO
  & 71.0 & 71.0 & 562 & 4.2 \\
ARPO
  & 69.7 & 72.0 & 377 & \textbf{3.3} \\
Random
  & 56.3 & 63.0 & 400 & 3.9 \\
\rowcolor{green!15}
\textbf{\ourmodel{} (ours)}
  & \textbf{75.8} & \textbf{78.0} & \textbf{344} & 3.6 \\
\midrule
\multicolumn{5}{l}{\textit{\textbf{Qwen2.5-3B-Instruct}}} \\
GRPO
  & 66.8 & 71.0 & 367 & \textbf{2.6} \\
DAPO
  & 72.7 & 73.0 & 533 & 2.7 \\
ARPO
  & 68.7 & 72.0 & 497 & 3.5 \\
Tree-Random
  & 72.7 & 75.0 & 419 & 3.6 \\
\rowcolor{green!15}
\textbf{\ourmodel{} (ours)}
  & \textbf{74.3} & \textbf{79.0} & \textbf{327} & 2.8 \\
\bottomrule
\end{tabular}
}
\caption{FrozenLake results across two model scales. We report success rate, Pass@4, and average response length (RLen) for each rollout method.}
\label{tab:frozenlake}
\end{table}

\begin{figure*}[t]
    \centering
    \includegraphics[width=\linewidth]{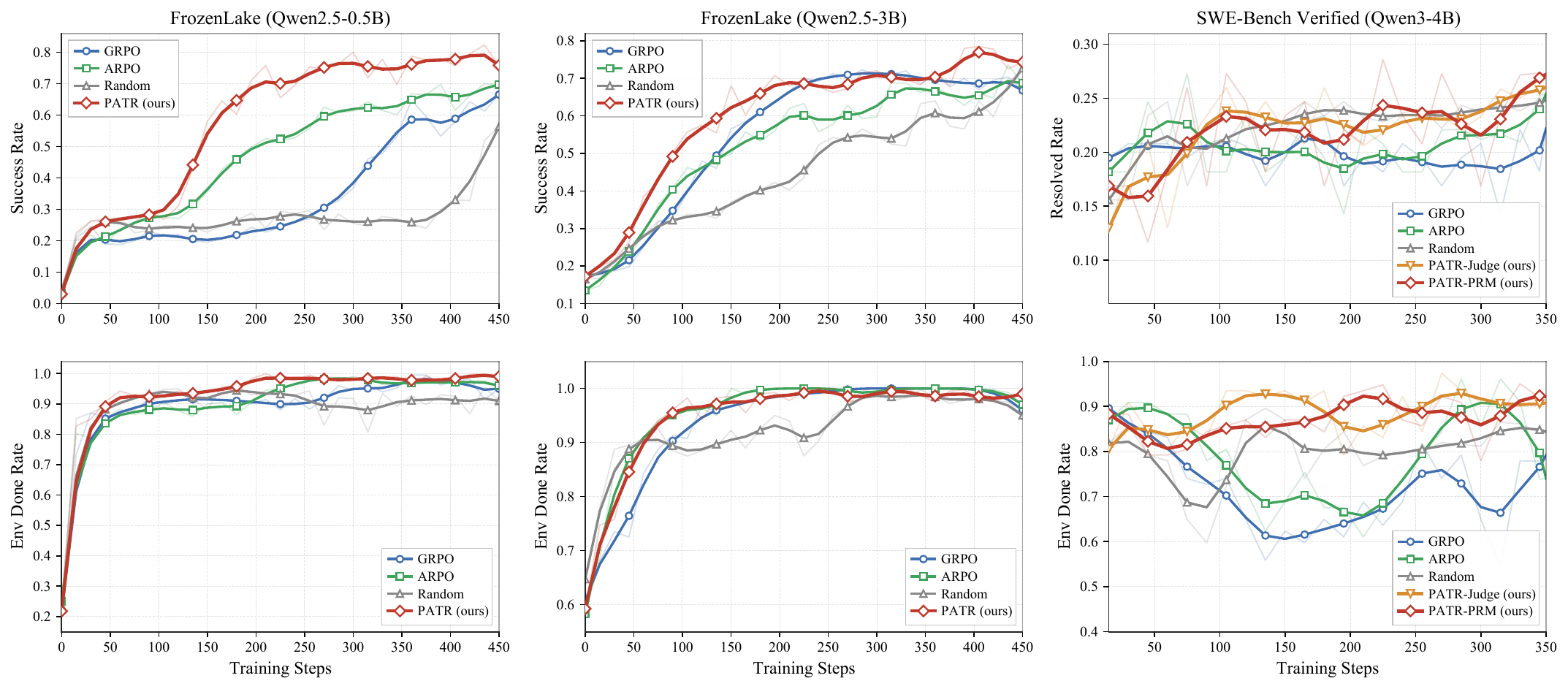}
    \caption{Training curves on FrozenLake and SWE-Bench across methods. \ourmodel{} improves performance earlier in training and maintains stronger trajectory-completion behavior throughout optimization.}
    \label{fig:train-curve}
    \vspace{-2mm}
\end{figure*}

\begin{table}[t]
\centering
\small
\setlength{\tabcolsep}{8pt}
\renewcommand{\arraystretch}{1.15}
\resizebox{\linewidth}{!}{
\begin{tabular}{lccc}
\toprule
\textbf{Method}
  & \textbf{Resolved Rate} $\uparrow$
  & \textbf{Env Done} $\uparrow$
  & \textbf{Turns} $\downarrow$ \\
\midrule
GRPO
  & 22.2 & 79.2 & 25.8 \\
DAPO
  & 24.6 & 71.4 & 31.6 \\
ARPO
  & 25.4 & 74.0 & 27.3 \\
Tree-Random
  & 24.8 & 84.4 & 30.1 \\
\rowcolor{green!10}
\textbf{\ourmodel-Judge (ours)}
  & 26.0 & 90.8 & \textbf{24.1} \\
\rowcolor{green!20}
\textbf{\ourmodel-PRM (ours)}
  & \textbf{27.2} & \textbf{92.2} & 24.4 \\
\bottomrule
\end{tabular}
}
\caption{SWE-Bench results with Qwen3-4B-Instruct-2507. We compare methods using resolved rate, environment completion rate (Env Done), and average turns.}
\label{tab:swe}
\vspace{-4mm}
\end{table}

\subsection{Main Results}
\label{sec:results-main}

\paragraph{FrozenLake.}
We report the final FrozenLake performance across two model scales in Table~\ref{tab:frozenlake}.
\ourmodel{} achieves the highest success rate and Pass@4 for both Qwen2.5-0.5B-Instruct and Qwen2.5-3B-Instruct.
The gain is especially pronounced for the smaller model, where \ourmodel{} improves the success rate from $66.5\%$ with GRPO to $75.8\%$, suggesting that process-guided rollout allocation is particularly useful when the base policy is less reliable.
Beyond accuracy, \ourmodel{} also produces the shortest average response length on both model scales, indicating that the gains do not come from simply exploring longer trajectories.
The comparison with Tree-Random further isolates the effect of process guidance.
Although Tree-Random benefits from shared-prefix rollout construction and is competitive in some settings, it remains below \ourmodel{} in both success rate and Pass@4.
This shows that \ourmodel{} gains from combining tree-structured exploration with quality-aware expansion, rather than from tree construction alone.
DAPO also improves over GRPO but often produces longer trajectories, highlighting the difference between objective-level improvements and rollout-level allocation.

\paragraph{SWE-Bench.}
We next evaluate the more challenging SWE-Bench Verified setting in Table~\ref{tab:swe}.
Both \ourmodel{} variants outperform all baselines in the resolved rate.
\ourmodel-PRM achieves the best result, improving over GRPO by $+5.0$ points and over the strongest baseline ARPO by $+1.8$ points.
\ourmodel-Judge also improves over all baselines, showing that the framework can benefit from either pretrained PRM scoring or LLM-as-judge feedback.
Both \ourmodel-PRM and \ourmodel-Judge complete substantially more task instances than the baselines while using fewer turns on average than DAPO, ARPO, and Tree-Random.
This suggests that process-guided expansion helps the policy learn more directed interaction patterns that reach natural task completion.
Tree-Random achieves a high environment-done rate but lower resolved rate, indicating that tree rollouts alone can improve completion behavior, while process-guided branch selection is needed to better align exploration with successful issue resolution.
The stronger performance of \ourmodel-PRM suggests that step-level PRM scores provide more stable branch-ranking signals for SWE-style code-editing tasks, while the judge variant remains a flexible alternative when task-specific PRMs are unavailable.

\begin{figure*}[t]
    \centering
    \includegraphics[width=\linewidth]{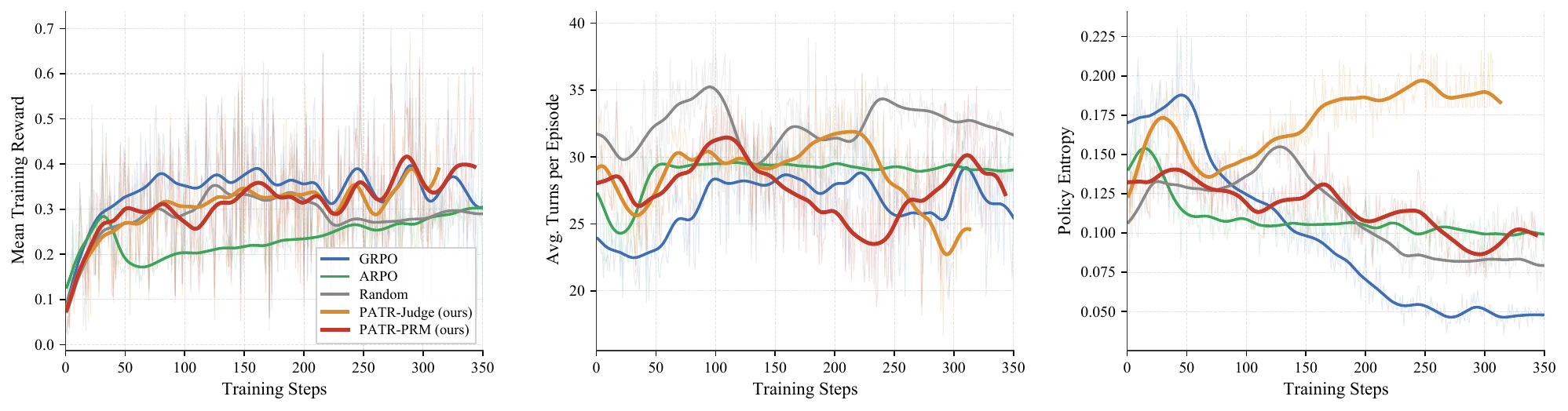}
    \caption{Training dynamics on SWE-Bench. \ourmodel{} improves the balance between reward acquisition, interaction length, and policy entropy compared with flat and unguided tree-rollout baselines.}
    \label{fig:analysis-dynamics}
    \vspace{-2mm}
\end{figure*}

\subsection{Training Dynamics}
\label{sec:results-dynamics}

We further analyze training dynamics in Figure~\ref{fig:train-curve}.
\ourmodel{} reaches higher success rates earlier than the baselines on FrozenLake, especially with the smaller Qwen2.5-0.5B model.
This supports the hypothesis that process-guided branching improves sample efficiency by focusing exploration around partial trajectories that already show progress.
The larger gap in the low-capacity setting further suggests that adaptive rollout construction can compensate for weaker initial policies by providing more informative training groups.

On SWE-Bench, \ourmodel-PRM and \ourmodel-Judge maintain higher resolved rates and environment-done rates across training.
In contrast, flat rollout methods improve more slowly or plateau earlier, while tree-based baselines without process-guided selection show less consistent gains.
The training curves therefore reinforce the main results: shared-prefix tree construction improves rollout diversity, but quality-aware expansion is important for converting additional exploration into reliable task progress.
Overall, the dynamics indicate that \ourmodel{} improves not only final performance but also the efficiency and stability of multi-turn agent RL.
\section{Analysis}
\label{sec:analysis}

\subsection{Effect of Tree Hyperparameters}
\label{sec:analysis-hyper}

We examine how tree construction choices affect \ourmodel-PRM on SWE-Bench Verified.
As shown in Table~\ref{tab:ablation}, the default setting, with branch factor $M{=}2$, scoring interval $K{=}13$, and top-$k{=}2$, achieves the best resolved rate.
Expanding more branches at each checkpoint slightly reduces performance, suggesting that overly broad expansion weakens the process-score filter.
Increasing the branch factor to $M{=}3$ also hurts performance, especially when expansion is concentrated on a single selected branch, indicating reduced rollout-group diversity.
Longer scoring intervals further degrade performance, showing that delayed rescoring limits the ability to redirect exploration during long-horizon interaction.
Overall, \ourmodel{} benefits from a moderate branching strategy that scores frequently while keeping expansion selective.

\begin{table}[t]
\centering
\small
\setlength{\tabcolsep}{7pt}
\renewcommand{\arraystretch}{1.15}
\resizebox{\linewidth}{!}{
\begin{tabular}{ccc|c}
\toprule
\textbf{Branch Factor} ($M$)
  & \textbf{Interval} ($K$)
  & \textbf{Top-}$k$
  & \textbf{Resolved Rate} $\uparrow$ \\
\midrule
\rowcolor{gray!12} \multicolumn{4}{l}{\textit{Varying Top-$k$}} \\
2 & 13 & 4 & 26.6 \\
\midrule
\rowcolor{gray!12} \multicolumn{4}{l}{\textit{Varying Branch Factor}} \\
3 & 13 & 2 & 25.8 \\
3 & 13 & 1 & 22.0 \\
\midrule
\rowcolor{gray!12} \multicolumn{4}{l}{\textit{Varying Branch Interval}} \\
2 & 15 & 2 & 23.4 \\
2 & 20 & 2 & 24.6 \\
\midrule
\rowcolor{green!20} \multicolumn{4}{l}{\textit{\textbf{\ourmodel-PRM (default)}}} \\
\rowcolor{green!20} \textbf{2} & \textbf{13} & \textbf{2} & \textbf{27.2} \\
\bottomrule
\end{tabular}
}
\caption{Ablation of tree rollout hyperparameters on SWE-Bench Verified. We vary the branch factor $M$, scoring interval $K$, and number of expanded branches $k$ per checkpoint.}
\label{tab:ablation}
\vspace{-3mm}
\end{table}

\subsection{Effect of Rollout Group Construction}
\label{sec:analysis-keepall}

We study whether pruned trajectories should be retained in the rollout group for GRPO training.
Table~\ref{tab:keepall} compares \textit{Best\_n}, which excludes pruned trajectories, with \textit{Keep All}, which retains them together with completed branches.
Keeping all trajectories consistently improves both \ourmodel-PRM and \ourmodel-Judge, with a larger gain for \ourmodel-Judge from $23.4\%$ to $26.0\%$ resolved rate.
This supports our design choice of preserving the full tree output for GRPO.
Because GRPO computes advantages relative to trajectories from the same task, failed and early-pruned branches can provide useful reward contrast rather than being discarded.
The lower average turns under \textit{Keep All} further suggests that pruned trajectories serve as informative negative examples without requiring full interaction budgets.

\begin{table}[t]
\centering
\small
\setlength{\tabcolsep}{8pt}
\renewcommand{\arraystretch}{1.15}
\resizebox{\linewidth}{!}{
\begin{tabular}{ll|cc}
\toprule
\textbf{Method} & \textbf{Rollout Group} & \textbf{Resolved Rate} $\uparrow$ & \textbf{Turns} $\downarrow$ \\
\midrule
\multirow{2}{*}{\ourmodel-PRM}
  & Best\_n   & 25.8  & 26.4 \\
  & Keep All  & \textbf{27.2} & \textbf{24.4} \\
\midrule
\multirow{2}{*}{\ourmodel-Judge}
  & Best\_n   & 23.4  & 31.2 \\
  & Keep All  & \textbf{26.0} & \textbf{24.1} \\
\bottomrule
\end{tabular}
}
\caption{Effect of rollout group construction on SWE-Bench Verified. \textit{Keep All} retains pruned trajectories in the GRPO group, while \textit{Best\_n} excludes them.}
\label{tab:keepall}
\vspace{-4mm}
\end{table}

\subsection{Training Signal and Policy Dynamics}
\label{sec:analysis-dynamics}

Figure~\ref{fig:analysis-dynamics} compares training reward, interaction length, and policy entropy on SWE-Bench.
Both \ourmodel{} variants obtain competitive training rewards while avoiding the excessive interaction length observed in some baselines.
\ourmodel-PRM maintains strong reward trends with relatively short episodes, consistent with its higher resolved rate in Table~\ref{tab:swe}.
The entropy curves show different exploration patterns: flat GRPO rapidly loses entropy, while \ourmodel-Judge maintains higher entropy for longer and \ourmodel-PRM follows a more conservative profile.
These dynamics suggest that process-guided tree rollout improves the balance between exploration, trajectory length, and reward acquisition during training.

\section{Conclusion}
We introduced \ourmodel, a process-guided adaptive tree rollout framework for multi-turn agent RL.
Instead of sampling complete trajectories independently, \ourmodel organizes rollout groups as trees and uses task-adaptive process feedback to decide where to expand, preserve, or prune partial trajectories.
The resulting rollout groups remain compatible with standard GRPO, since process scores guide rollout construction while policy optimization still relies on task outcome rewards.
Experiments on FrozenLake and SWE-Bench show consistent gains over independent-rollout and other rollout-generation baselines, with improvements of up to $+5.0$ points on SWE-Bench and $+9.3$ points on FrozenLake.
These results highlight process-guided tree rollouts as a promising direction for effective and scalable multi-turn agent learning.

\section*{Limitations}

This work has several limitations.
First, \ourmodel{} relies on the quality of the process scorer used to rank partial trajectories.
Although we instantiate the scorer with heuristics, pretrained PRMs, and LLM-as-judge feedback, inaccurate process scores may lead the tree rollout to expand suboptimal branches or prune useful ones.
In particular, the pretrained PRM used in our SWE-Bench experiments, Skywork-o1-Open-PRM-Qwen-2.5-7B, may favor actions that generate longer responses, which can affect branch ranking and rollout allocation.
Second, our experiments focus on FrozenLake and SWE-Bench.
While these benchmarks cover both controlled navigation and realistic long-horizon coding-agent tasks, further evaluation is needed on broader multi-turn agent domains, such as web navigation, embodied interaction, and open-ended tool use.

\paragraph{LLM Usage Disclosure}
LLMs were used only for grammar correction and writing polishing.

\bibliography{custom}

\appendix
\section{Additional Implementation Details}
\label{app:impl-details}

\paragraph{Training details.}
We use the same optimization hyperparameters across methods unless otherwise specified.
The learning rate is $1\times10^{-6}$, the clipping ratio is $0.28$, the KL coefficient is $0.001$, and the rollout sampling temperature is $1.0$.
FrozenLake is trained with batch size 64 for 400 steps, while SWE-agent training uses batch size 8 for 300 steps.
FrozenLake experiments are run on one A100 node, and SWE-agent experiments are run on one H200 node.

\paragraph{Process scoring details.}
For FrozenLake, the process scorer combines accumulated step rewards with a normalized BFS shortest-path distance to the goal, where holes are treated as blocked cells.
For SWE-agent training, both \ourmodel-PRM and \ourmodel-Judge score each branch using recent trajectory context.
\ourmodel-PRM serializes the most recent 10 interaction steps and feeds them to Skywork-o1-Open-PRM-Qwen-2.5-7B, which returns step-level process scores; we use the score of the most recent step as the branch score.
\ourmodel-Judge provides Qwen2.5-Coder-7B-Instruct with the task context and a compact representation of recent actions, observations, tool types, and extracted trajectory signals.
The judge outputs a trajectory-level score, which is blended with a lightweight heuristic that rewards useful signals such as targeted testing and evidence-based editing while penalizing repeated actions and premature termination.
We describe the heuristic and provide all judge prompts in Appendix~\ref{app:judge-prompts}.

\paragraph{Scoring infrastructure.}
For SWE-bench experiments, the PRM and LLM-judge models are hosted on separate vLLM inference servers and queried during expansion checkpoints.
This separates policy rollout generation from process scoring and avoids GPU contention during training.

\section{LLM-as-Judge Scoring Details}
\label{app:judge-prompts}

This section provides additional details about the LLM-as-judge scorer used in \ourmodel-Judge for SWE-Bench.
At each scoring checkpoint, the judge receives the task description and a compact representation of the current partial trajectory, including recent agent actions, environment observations, inferred tool types, and automatically extracted trajectory signals.
The judge outputs a scalar score in $[0,1]$ indicating how promising the current branch is for further expansion.
To make the evaluation more stage-aware, we use different prompts for diagnosis, editing, and verification phases.

\paragraph{Phase selection.}
The prompt is selected automatically from recent interaction patterns.
If the recent trajectory contains test-related actions or observations, such as running \texttt{pytest}, \texttt{unittest}, or observing pass/fail messages, we use the verification prompt.
If the recent trajectory contains file modification actions, such as patching, insertion, or replacement, we use the editing prompt.
If the recent trajectory mainly contains repository exploration, such as search, file inspection, traceback analysis, or command-line reading, we use the diagnosis prompt.
When no specific phase is detected, we use the default prompt.
This design allows the judge to evaluate the type of progress that is most relevant to the current stage of problem solving.

\paragraph{Heuristic score.}
In addition to the LLM score, we compute a lightweight heuristic score from recent trajectory signals.
The heuristic starts from a neutral value and adjusts the score according to observable behaviors.
It penalizes repeated identical actions, premature finish actions, command failures, and edits made without prior evidence from search, file inspection, or testing.
It rewards signals associated with useful progress, including targeted test execution, passing tests, informative tracebacks, code or file information discovered during exploration, and edits made after relevant evidence has been collected.
The final branch score is a weighted combination of the LLM judgment and the heuristic score, with the LLM judgment receiving the larger weight.
The heuristic acts as a stabilizing signal, especially when the LLM judge assigns overly optimistic scores to repetitive or weakly grounded trajectories.

\paragraph{Judge prompts.}
We list the system prompt and all phase-specific scoring prompts below~\ref{lst:judge-system}, ~\ref{lst:judge-default},~\ref{lst:judge-diagnose},~\ref{lst:judge-edit}, and~\ref{lst:judge-verify}.
The placeholders \texttt{\{task\}} and \texttt{\{trajectory\}} are filled with the task description and the compact trajectory representation at each scoring checkpoint.

\begin{figure*}[t]
\begin{minipage}{1.0\textwidth}
\begin{lstlisting}[
basicstyle=\ttfamily\footnotesize,
breaklines=true,
caption={System prompt for the LLM-as-judge scorer.},
label={lst:judge-system}
]
You evaluate partial SWE-agent trajectories for tree-search expansion.
Output ONLY valid JSON: {"score": <0.0 to 1.0>}. Nothing else.
\end{lstlisting}
\end{minipage}
\end{figure*}

\begin{figure*}[t]
\begin{minipage}{1.0\textwidth}
\begin{lstlisting}[
basicstyle=\ttfamily\footnotesize,
breaklines=true,
caption={Default judge prompt.},
label={lst:judge-default}
]
Score this partial SWE-agent trajectory for tree-search branching.

Consider:
- Is the agent making measurable progress toward solving the issue?
- Are its actions grounded in evidence from the code or observations?
- Is it avoiding repetitive or aimless commands?
- If we continue from this state, how likely will this branch lead to a correct fix?

0.0 = stuck, looping, or wrong direction
0.5 = some relevant work but unclear progress
1.0 = clearly on track, strong evidence of progress

Task:
{task}

Trajectory:
{trajectory}

Output ONLY: {"score": <0.0 to 1.0>}
\end{lstlisting}
\end{minipage}
\end{figure*}

\begin{figure*}[t]
\begin{minipage}{1.0\textwidth}
\begin{lstlisting}[
basicstyle=\ttfamily\footnotesize,
breaklines=true,
caption={Judge prompt for the diagnosis phase.},
label={lst:judge-diagnose}
]
Score this partial trajectory during EXPLORATION / DIAGNOSIS.

Consider:
- Is the agent searching in the right files and directories?
- Is it narrowing down the bug location with each step?
- Is it gathering evidence (reading code, reproducing the bug) rather than guessing?
- Is it avoiding repeated searches that yield no new information?

0.0 = aimless browsing, wrong files, repetitive commands
0.5 = relevant exploration but slow progress
1.0 = efficiently locating the root cause

Task:
{task}

Trajectory:
{trajectory}

Output ONLY: {"score": <0.0 to 1.0>}
\end{lstlisting}
\end{minipage}
\end{figure*}

\begin{figure*}[t]
\begin{minipage}{1.0\textwidth}
\begin{lstlisting}[
basicstyle=\ttfamily\footnotesize,
breaklines=true,
caption={Judge prompt for the editing phase.},
label={lst:judge-edit}
]
Score this partial trajectory during EDITING.

Consider:
- Is the agent editing the file that evidence pointed to?
- Is the change minimal and tied to the diagnosed root cause?
- Did the agent collect enough evidence before editing, or is it guessing?
- Could this edit introduce new bugs or break other functionality?

0.0 = editing blind, wrong file, speculative changes
0.5 = reasonable edit but unclear if it addresses root cause
1.0 = precise, evidence-based fix in the right location

Task:
{task}

Trajectory:
{trajectory}

Output ONLY: {"score": <0.0 to 1.0>}
\end{lstlisting}
\end{minipage}
\end{figure*}

\begin{figure*}[t]
\begin{minipage}{1.0\textwidth}
\begin{lstlisting}[
basicstyle=\ttfamily\footnotesize,
breaklines=true,
caption={Judge prompt for the verification phase.},
label={lst:judge-verify}
]
Score this partial trajectory during TESTING / VERIFICATION.

Consider:
- Is the agent running tests relevant to the issue (not random test suites)?
- If tests fail, does the agent appear to understand the failure?
- If tests pass, do they actually cover the bug that was fixed?
- Is the agent verifying the fix rather than declaring premature success?

0.0 = irrelevant tests, ignoring failures, false success
0.5 = running some tests but coverage of the fix is unclear
1.0 = targeted verification that confirms the fix works

Task:
{task}

Trajectory:
{trajectory}

Output ONLY: {"score": <0.0 to 1.0>}
\end{lstlisting}
\end{minipage}
\end{figure*}

\end{document}